Scientific
Research
Publishing

# An Improved Real-Time Face Recognition System at Low Resolution Based on Local Binary Pattern Histogram Algorithm and CLAHE

## Kamal Chandra Paul[1], Semih Aslan[2]

[1]Electrical and Computer Engineering, The University of North Carolina at Charlotte, North Carolina, USA
[2]Ingram School Engineering, Texas State University, San Marcos, TX, USA
Email: kpaul9@uncc.edu, sa40@txstate.edu

## Abstract

This research presents an improved real-time face recognition system at a low resolution of 15 pixels with pose and emotion and resolution variations. We have designed our datasets named LRD200 and LRD100, which have been used for training and classification. The face detection part uses the Viola-Jones algorithm, and the face recognition part receives the face image from the face detection part to process it using the Local Binary Pattern Histogram (LBPH) algorithm with preprocessing using contrast limited adaptive histogram equalization (CLAHE) and face alignment. The face database in this system can be updated via our custom-built standalone android app and automatic restarting of the training and recognition process with an updated database. Using our proposed algorithm, a real-time face recognition accuracy of 78.40% at 15 px and 98.05% at 45 px have been achieved using the LRD200 database containing 200 images per person. With 100 images per person in the database (LRD100) the achieved accuracies are 60.60% at 15 px and 95% at 45 px respectively. A facial deflection of about 30° on either side from the front face showed an average face recognition precision of 72.25% - 81.85%. This face recognition system can be employed for law enforcement purposes, where the surveillance camera captures a low-resolution image because of the distance of a person from the camera. It can also be used as a surveillance system in airports, bus stations, etc., to reduce the risk of possible criminal threats.

## Keywords

Face Detection, Face Recognition, Low Resolution, Feature Extraction,

## 1. Introduction

The process of detecting and locating faces from a single or series of images and identifying the face(s) is known as face recognition. As humans, we are very good at detecting and recognizing faces; however, it's difficult for computers to detect and recognize faces. Face recognition has diversified applications, including video surveillance systems, security, access control, law enforcement, general identity verification [1], gender recognition [2], missing person identification, etc. While implementing these applications, face detection comes before recognition. Face recognition can be classified into two categories, namely, face verification and face identification. Face verification refers to the process which can detect whether a pair of pictures belongs to the same individual or not.

On the other hand, face identification refers to the labeling target face with respect to a training set or database. Face recognition can be done using 1) holistic matching method, or 2) structural method, or 3) hybrid method [3]. In a holistic approach (e.g., Eigenfaces, PCA [4] [5], or Linear Discriminant Analysis [6]), the entire face image is considered for face recognition. Whereas, in structural methods, essential features like nose, mouth, eyes, and their locations and native statistics are considered for recognition purposes. Hybrid methods utilize a combination of holistic and structural methods. Using a face recognition system, an input face image is compared to those face images recorded previously and kept in the system's training database set.

During the past few decades, researchers have introduced various face detection and face recognition methods. The Viola-Jones algorithm [7] is one of the most popular face detection algorithms, and it is the basis of many other face detection techniques. In this algorithm, Haar-like rectangle features are used to detect a face. They have used an image size of $320 \times 240$ pixels and got comparatively inferior performance with respect to the software-based system executed on CPUs. Chakrasali and Kuthale [8] used Haar features and the AdaBoost algorithm for face detection.

In the face recognition process, face detection plays a vital role. In some recognition systems, holistic face recognition processes have been used. Principal Component Neural Network (PCNN) based face recognition has been implemented in [9]. This system can recognize up to 1400 faces in an image frame suitable for access control and video surveillance. They have presented a real-time face recognition system with an image resolution of $32 \times 32$ pixels only. Reference [10] supports a low-resolution image of $60 \times 60$ pixels, but a real-time application is not specified. Endluri *et al.* [11] reported a real-time embedded face recognition system based on the PCA method, which supports resolution $320 \times 340$ pixels' images. Also, this system can't be used for more than two users. Recently, Schaffer *et al.* presented [12] a face detection system that utilized a software-based (MATLAB) Viola-Jones face detection algorithm and FPGA based PCA algorithm for the face recognition part. They have reported real-time face recognition at an accuracy of about 95% and can process 13,026 faces per second. In this system, 20 face images from 153 people have been considered for the recognition database [13] [14]. Zhao and Wei [15] presented MLBPH based real-time face recognition system with various facial deflection and attitudes and achieved recognition accuracy of about 48% - 55% for 30° angular positions of the face. Ahmed *et al.* [16] also used LBPH architecture for face recognition at a low resolution of 35 px. But they have used 500 images per person and reported recognition efficiency of 94% at 45 px and 90% at 35 px. [17] proposed an LBPH based face recognition on GPU. Our proposed method employs the LBPH algorithm along with face alignment and CLAHE for real-time face recognition at low resolution with higher recognition accuracy. It has an improved face recognition performance with a reduced number of images person in the database. Our experimental results indicate that face recognition can achieve an improved result at only 15 px of the input image even with various attitudes and facial deflection.

## 2. Overall Proposed Design

### 2.1. Face Detection

This section presents an overview of the face detection technique, which employs the Viola-Jones [7] algorithm. To construct the classifiers, the Viola-Jones algorithm uses Haar-like rectangle features, as shown in **Figure 1**. For example, Haar-like rectangle c in **Figure 1** is used to detect a human face's eyes (**Figure 2**) because the area covered by the eyes is darker than the area just above the cheeks. Haar feature f in **Figure 1** can be used to detect the nose feature because the nose's junction area is brighter with respect to its two chick sides (**Figure 2**). A crucial part of this algorithm is to compute the rectangle features very quickly to form an integral image. The integral image is constructed in such a way so that at the location $(x, y)$ of the integral image, it contains the sum of all the pixels to the left and above of it.

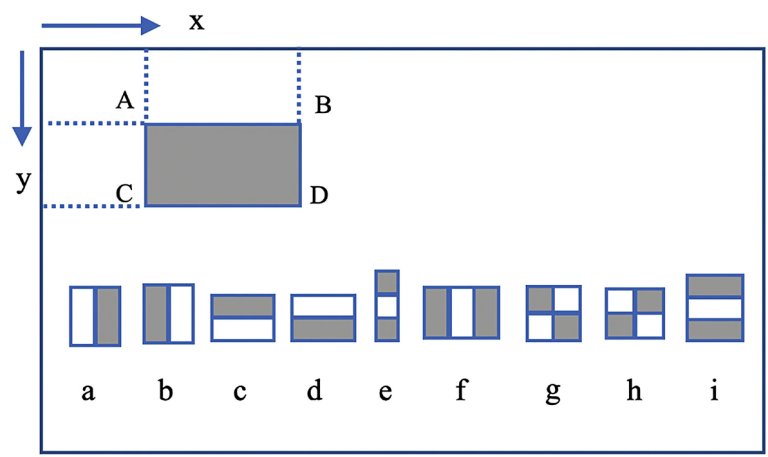

**Figure 1.** Block diagram of the system.

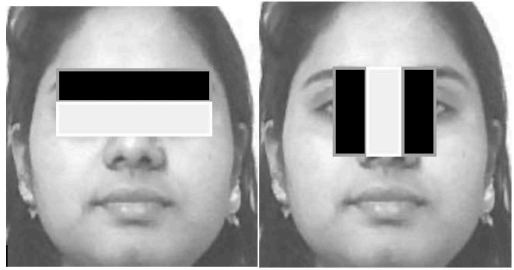

**Figure 2.** Relevant Haar feature for face detection.

A cascade classifier is used in this algorithm, which is constructed as a sequence of stages. At each stage, a set of selected rectangle features are used to slide over a sub-window (**Figure 3**) to check whether there is a face or not. A sub-window region is either rejected as a face candidate or is pushed to the next state for further processing using a threshold check. In order to detect faces of different sizes, the algorithm uses a pyramid of scaled images that consists of the same set of rectangle features but of different sizes to slide over the initial image until all faces are found. Finally, all the faces are marked with red rectangles in the original test image.

## 2.2. Feature Extraction Using LBPH Algorithm

Local Binary Pattern (LBP), a powerful feature for texture classification in computer vision, is a simple yet very efficient operator to describe a pixel's contrast information with respect to its neighboring pixels. It labels an image's pixels by thresholding the neighboring pixels and considers the result as a binary number. When combined with Histograms of Oriented Gradients (HOG) descriptor, LBP considerably improves performance on some datasets. LBP combined with HOG descriptor can represent a facial image as a simple data vector, and it can be used for face recognition purposes. In the Local Binary Pattern Histogram (LBPH) algorithm of face recognition shown in **Figure 4**, the face image is converted into a grayscale image.

For feature extraction, the grayscale image is divided into 3 × 3 window cells (**Figure 5**), and the center pixel of the cell is compared with each of the surrounding 8 pixels either clockwise or in a counterclockwise direction. If the surrounding pixel is greater than the center pixel, it is replaced with one; otherwise, it is replaced with zero. Thus, if we start counting in a clockwise manner in the resultant 3 × 3 window except for the center value, we get a binary number of 8 bit and is replaced with the decimal equivalent of the binary number with the center pixel in the original cell. This value is used to reflect the texture feature of that region.

Let $g_c$ and $g_0, g_1, g_2, \ldots, g_{P-1}$ denote the values of center pixel and neighbor pixels, respectively, then we can get the 8 bit LBP code with respect to the center pixel at position $(x, y)$ using Equation (1).

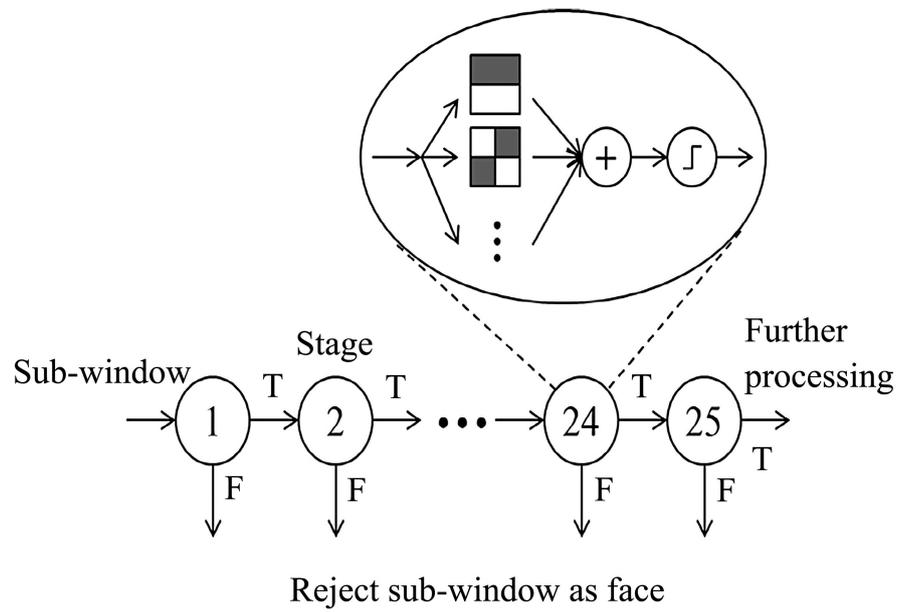

**Figure 3.** A cascade classifier of 25 stages & decision tree (T = Ture, F = False) [11].

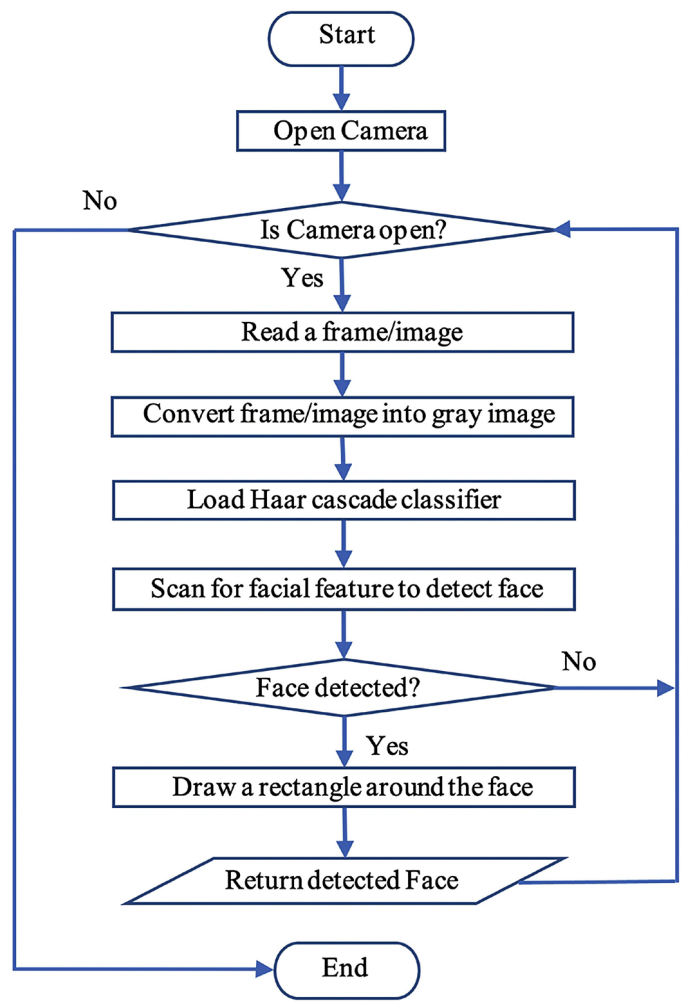

**Figure 4.** Flowchart for face detection.

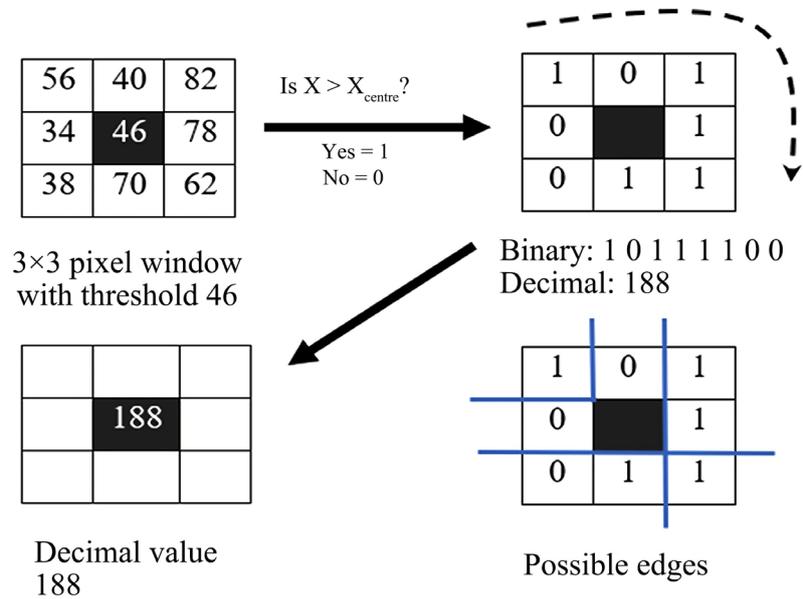

**Figure 5.** A 3 × 3 LBP operator.

$$\text{LBP}\left(x_c, y_c\right) = \sum_{p=0}^{p-1} S\left(g_c - g_p\right) 2^p \tag{1}$$

The threshold function $s(z)$ can be given by Equation (2)

$$s\left(z\right) = \begin{cases} 1, & z \geq 0 \\ 0, & z < 0 \end{cases} \tag{2}$$

In the LBPH algorithm, the histogram which is used as a texture descriptor is basically a collection of the LBP codes of all the pixels for an input image, *i.e.*,

$$\text{BPH}\left(i\right) = \sum_{x_c, y_c} \delta\left\{i, \text{LBP}\left(x_c, y_c\right)\right\}, i = 0, 1, \cdots, 2^{p-1} \tag{3}$$

where $\delta(.)$ is known as the Kroneck product function.

The LBPH method allows us to make the LBP operator of different radius and neighborhoods, known as circular LBP operator. **Figure 6** shows an example of an LBP operator with various number neighbors and radii, where $P$ denotes the number of neighboring pixels, and R denotes the radius of the circular LBP operator.

The whole gray face image is subdivided into several sub-regions in the LBPH algorithm, and then the LBP feature vectors of each sub-region are extracted. After that, a histogram of each sub-region is calculated from the LBP feature vectors. All the histograms of the sub-regions are then concatenated to get the bigger size histogram of the full image, which represents the main characteristic of the image.

## 2.3. Creating the Face Database and Training

For this study, we have created our own database of 5 people. We have created two sets of databases. One database named LRD200 shown in **Figure 7**, consisting of 200 face image per person, have a total of 1000 face images and another database named LRD100 consists of 100 face image per person totaling 500 face images in this database. Face database LRD200 contains all the face images of the LRD100 database. Hoping to get improved and more generic recognition, the images were captured in two different illumination conditions.

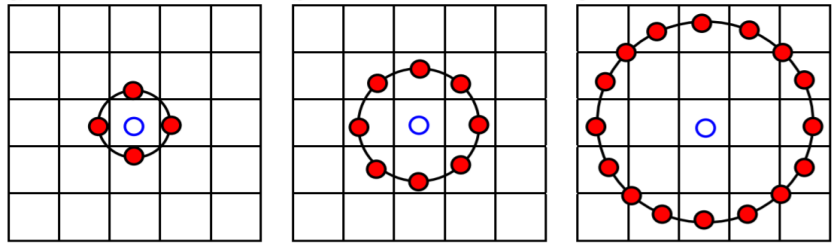

**Figure 6.** Circular neighborhoods of the center pixel with different neighbor pixels.

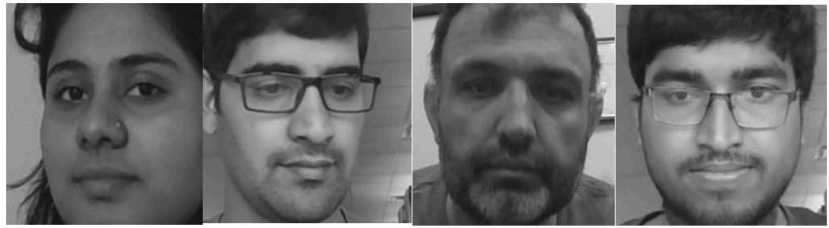

**Figure 7.** Sample face image of LRD200 database.

To create this database, we have built an Android App named MyNewCam. A few snapshots, along with the development details, are given in **Figure 8** and **Figure 9**. The app is developed in Android Studio using Java language. The compiled SDK version is 28 with camera API 23. While taking pictures of the subject face, this app requires the user to put the name of the person. This name, later, is used for recognition purposes during the face recognition stage. The subject images from the android app are then sent to a folder of the system computer, where we run a directory watcher program that detects any incoming face image. The directory watcher program, along with the face detector program, detects new images in the folder and then detects faces and aligns the face images in a vertical position so that the eyes are always in a horizontal position. To get better accuracy of face recognition, the face images are then preprocessed before cropping. We have applied median filtering to remove the noise present in the image. Afterward, the cropped face images along with the subject names are automatically saved to the "training-image" folder, which contains all the database images. This process also includes the automatic starting of the training process when new face images are added to the training-image folder. After completing the training process, the training data is then saved in the LBPHrecognizer_database.xml file, which is used for recognition purposes. A new face image can always be added using the android app, and the system then automatically trains the image database and reruns the face recognizer program. We have used OpenCV libraries and python for face detection and recognition. The process of creating a database and training is shown in **Figure 10**.

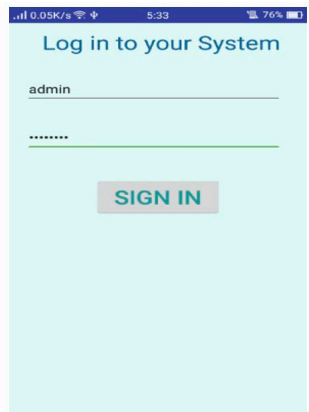

Camera app details:
**Compile SDK version**: 28,
**Camera API**: 23,
**Min SDK Version**: 18,
**Target SDK version**: 28,
**Platform**: Android Studio,
**Features**: Camera with person's name wise image capture, Image gallery to view all and single image along with zoom view.

**Figure 8.** Login screen of the camera app (left) and camera features (right).

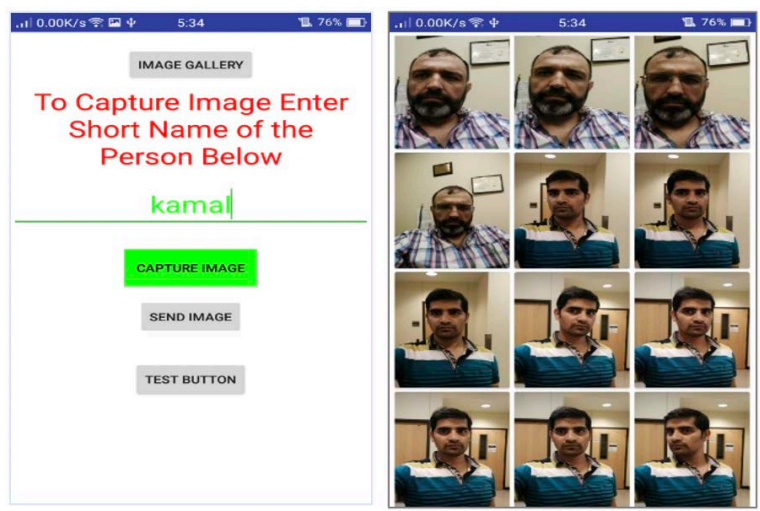

**Figure 9.** Capture button (left) & image gallery (right) of the Android App.

### 2.4. Face Recognition

We set up a real-time face recognition system where the input image was captured from the webcam video feed. Not every frame was used for recognition. Frames after every 100 milliseconds are counted for recognition. Before starting the webcam, the face database file for LBPH and NameList.txt files are loaded. Each frame read is passed through an image enhancement technique called CLAHE (Contrast Limited Adaptive Histogram Equalization) to enhance image quality, and then Gaussian filtering is applied to reduce the noise present in the image during capturing from the camera. The frame was then converted into a gray image for detecting a face. If the face image is found tilted, it is then aligned to keep both eyes in the same horizontal position. The detected face is cropped and resized. The LBP feature vectors are extracted, and a histogram of the face image is obtained using the LBPH algorithm, which represents the characteristic of the image.

To recognize the input image, its histogram is compared with the database histograms, and the image is recognized as the subject image in the database having the closest histogram. A threshold value is set to identify an unknown person who is not included in the image database. To compare histograms, we have used the Euclidean distance method.

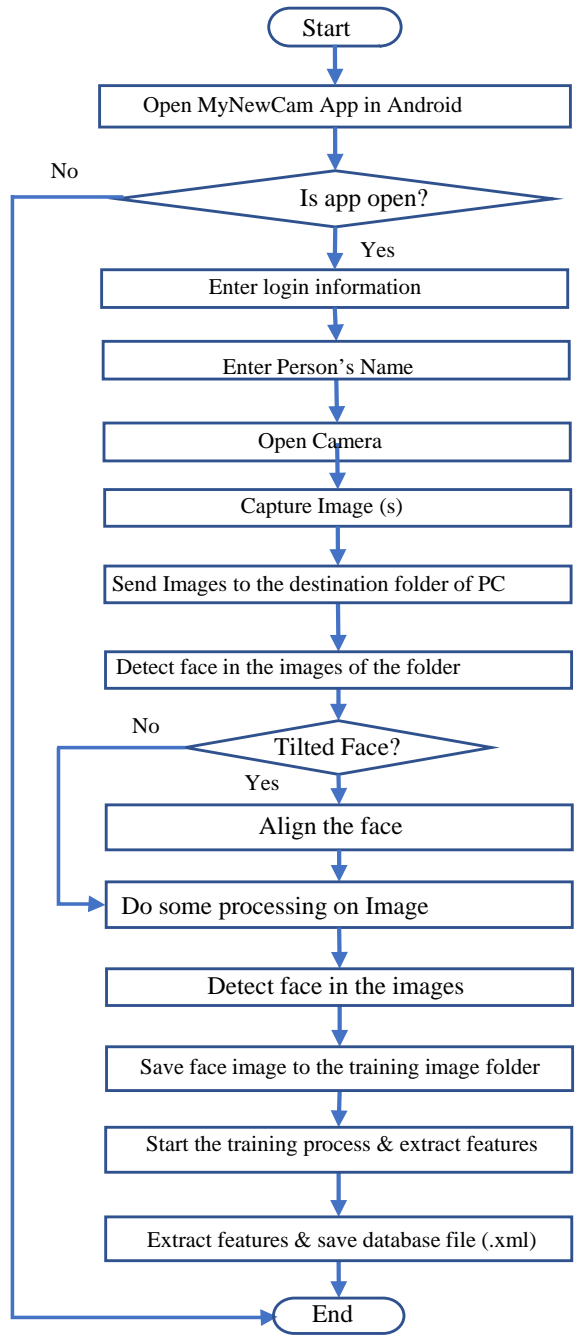

**Figure 10.** The process of creating a database and training.

To recognize face using OpenCV libraries and python, we need to have subject names and corresponding ID of the subjects. In this experiment, each of the database images has the subject name with an image file name and a "_" after the name field. To create the subject IDs, we have read all the images in the training-image folder and extract the name part from the image files, list only differ-ent subject names, and sort the names according to alphabetical order. After that, an ID is given with each subject name corresponding to the subject's position in the list. The subject names from the list and corresponding IDs of the subjects are then saved in a text file named "NameList.txt." When we get an ID of a subject name during the recognition process, the corresponding subject name is called from this text file.

For face recognition, we have taken the input images from the webcam of the laptop. We have experimented with the recognition rate with different low-resolution image frames. During the recognition process, image frames were taken from the webcam at an interval of 100 milliseconds. Each time 200 image frames were used for recognition, and the process was repeated ten times. Then the recognition rate is calculated out of those 2000 image frames. A complete flowchart of face recognition is depicted in Figure 11.

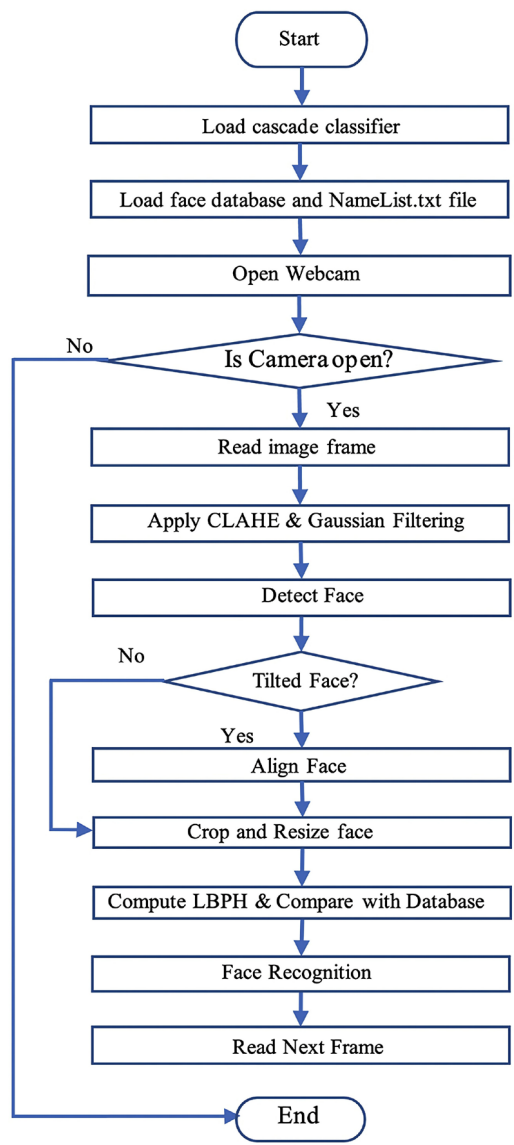

**Figure 11.** Flowchart for face recognition.

An overall proposed face recognition system is provided in Figure 12. The camera takes input video continuously and feeds real-time images to the recognition system, which is then recognized by the face recognition subsystem. A recognized face is marked with the person's name, and an unknown face is marked as unknown on display.

## 3. Experimental Results and Discussion

In our proposed face recognition system, we have used a core i5 Lenovo Think-Pad 14 with an integrated webcam of 0.9 MP to recognize real-time faces. For the database, we have used an android phone of model Oppo R7kf having an android version of 5.5.1. It has a back camera of 13 MP. To capture the images, no flash is used. In the experiment we have used low resolution images of 15 px, 20 px, 30 px, 35 px, and 45 px respectively to check the performance of real-time face recognition. The performances are determined using both LDR200 and LDR100 databases. A rotating head around the camera has been used for normal face recognition where the head is slowly moved from the left 30˚ to the right 30˚ with upward and downward faces.

### 3.1. Face Recognition under Different Low Resolutions

This section describes the result of face recognition for the different low resolution of the input images. The recognition rates, with rotating head around the camera, with our created database LRD200, are tabulated in Table 1, and those with LDR100 database are tabulated in Table 2. With the increase in image resolution, the recognition rate increases. Also, the number of images in the database plays a vital role in determining the recognition accuracy. Overall, 78.84% face recognition accuracy with the LDR200 database is achieved at only 15 pixels and 98.05% at 45 pixels of the input image, as shown in Table 3. The results indicate that with the increase of the input images' pixel, the recognition accuracy increases. The recognition accuracy is high even with a rotating head around the camera. This is because when the head is in a tilted position with respect to the camera, the cropped face image is aligned before being recognized, which helps to improve the recognition accuracy. On the other hand, CLAHE preprocesses the low-resolution images, which also helps in improving the face recognition performance. CLAHE improved the illumination variation within the images.

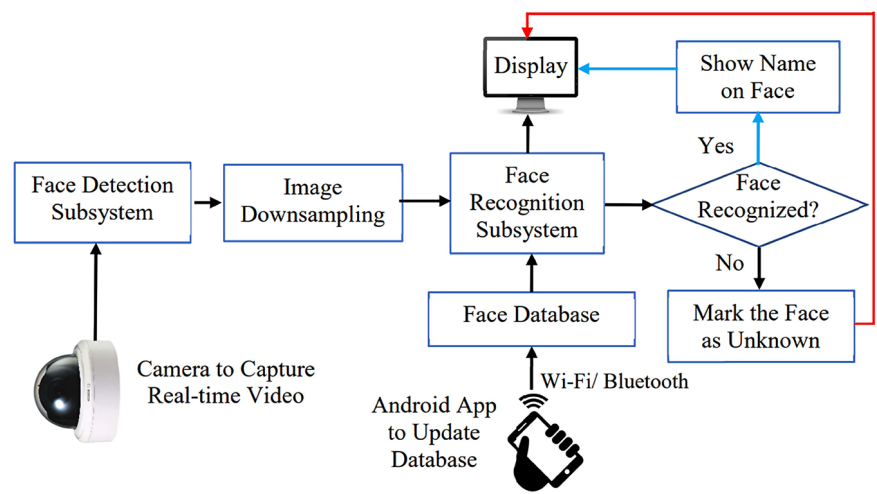

**Figure 12.** The proposed face recognition system.

**Table 1.** Recognition rate based on LRI200 database.

| Recognition | Using Database LRD200 | | |
|---|---|---|---|
| | *Correct Times* | *Wrong Times* | *Recognition Rate* |
| At 15 pixel | 1568 | 432 | 78.40% |
| At 20 pixel | 1842 | 158 | 92.10% |
| At 30 pixel | 1919 | 81 | 95.95% |
| At 35 pixel | 1932 | 68 | 96.60% |
| At 45 pixel | 1961 | 39 | 98.05% |

**Table 2.** Recognition rate based on LRD100 database.

| Recognition | Using Database LRD100 | | |
|---|---|---|---|
| | *Correct Times* | *Wrong Times* | *Recognition Rate* |
| At 15 pixel | 1212 | 642 | 60.60% |
| At 20 pixel | 1633 | 367 | 81.65% |
| At 30 pixel | 1691 | 309 | 84.55% |
| At 35 pixel | 1855 | 145 | 92.75% |
| At 45 pixel | 1900 | 100 | 95.00% |

**Table 3.** Recognition rate based on LRD200 database.

| Recognition at 45 px | Using Database LRD200 | | |
|---|---|---|---|
| | *Correct Times* | *Wrong Times* | *Recognition Rate* |
| Front Facing | 1993 | 7 | 99.65% |
| Facing 30° Right | 1637 | 383 | 81.85% |
| Facing 30° Left | 1545 | 455 | 72.25% |

Figure 13 shows the input face image under various resolution conditions while recognizing. Figure 14 shows a graphical representation of the face recognition rate for various low-resolution images. It also represents the trending of recognition accuracy with respect to the number of images per person in the database.

### 3.2. Face Recognition with Different Angular Positions

The recognition accuracy varies with the various angular positions of the head with respect to the camera. With a higher deflection angle, the recognition rate deteriorates. The recognition rates were recorded with three angular positions of

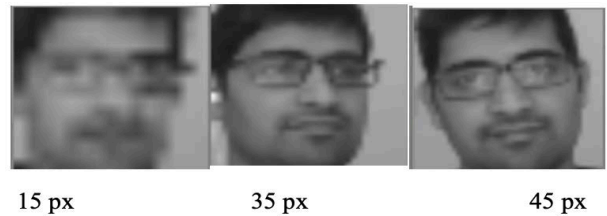

15 px      35 px      45 px

**Figure 13.** Face image to recognize at different resolution.

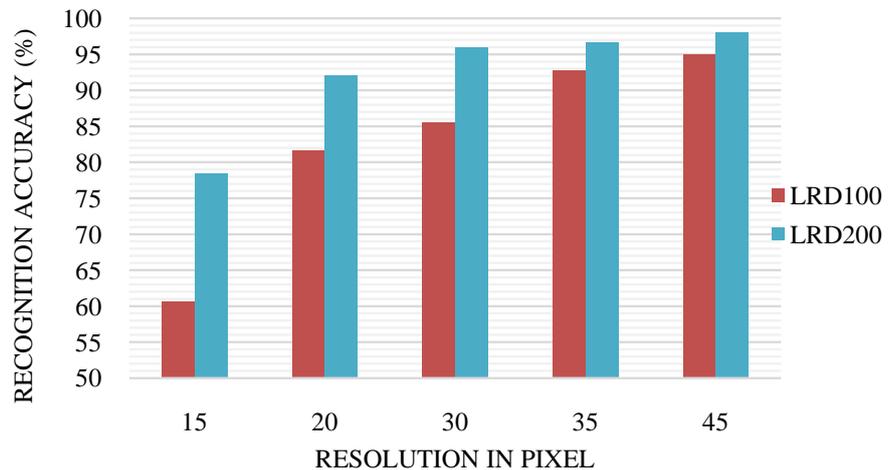

**Figure 14.** Graphical representation of recognition accuracy with image resolution (pixel).

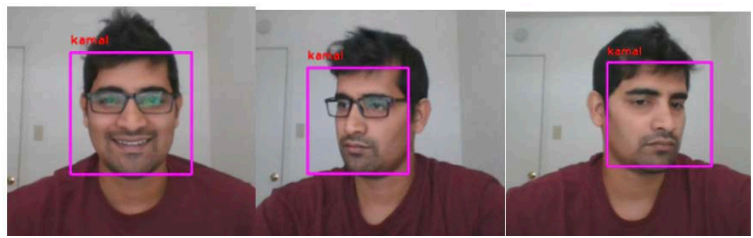

**Figure 15.** Face recognition with different angular deflection.

frontal head-front face, about 30° right, and about 30° left positions, as shown in **Figure 15**. While capturing image frames in the front face position, the head was moved up and downward direction. Similarly, the head was moved up and downward direction while the facial deflections were about 30° left and about 30° right positions. It has been found that the recognition rate was higher when the face is front faced only. Each time with 200 frames was used for recognition, and the process was repeated ten times to calculate the recognition accuracy out of 2000 frames.

## 4. Comparison of Our Results with Other Methods

This section proves a comparison of our proposed algorithm with prior methods. A detailed comparison of our proposed face recognition algorithm is provided in **Table 4**. Reference [16] showed face recognition at the lowest of 35 px with 90% accuracy, whereas our recognition system can recognizer real-time face images with an accuracy of 96.6% at 35 px. Moreover, our proposed method can recognize face images at as low as 15 px with a recognition accuracy of 78%. The number of images per person is also much lower in our database compared to reference [16]. With facial deflection, our proposed algorithm has a higher recognition performance compared to reference [15].

**Table 4.** Recognition rate based on LRD100 database.

| Particulars | Zhao and Wei [15] | Ahmed *et al.* [16] | Ours |
|---|---|---|---|
| Method | MLBPH | LBPH | LBPH + CLAHE + alignment |
| Lowest image resolution | - | 35px | 15px |
| Accuracy at 15 px | - | - | 78.4% |
| Accuracy at 35 px | - | 90% | 96.60% |
| Accuracy at 45 px | - | 94% | 98.05% |
| Accuracy with 30° angular deflection | 48.4% - 55% | - | 72.25% - 81.85% @ 45 px |
| Use of Android app for database | - | - | Yes |
| Auto-update of the database & auto-restart of process | No | No | Yes |
| Number of image per person in the database | 7 | 500 | 200 |

## 5. Conclusions

Local Binary Pattern Histogram (LBPH) architecture of face recognition is a powerful algorithm to recognize the face under varying illumination conditions and at low resolution. We have used median filtering for database images and Gaussian filtering for the recognition of a face. Our experiment results in a novel face recognition accuracy of 78.40% at the low resolution of 15 px and 98.05% at 45 px with LRD200 database. Also, with LRD100 data-based we have achieved 60.60% and 95% at 15 px and 45 px respectively. The recognition accuracy with attitude deflection showed an improved result. Moreover, an android app has been developed for capturing the subject image and sending images from the android app to the pc, which will automate the process of retraining and restarting the recognition process. This research also found that the number of images per person in the training set database plays a significant role in recognition accuracy at low resolution. A larger dataset having more diverse pose and illumination variation may increase the face recognition accuracy [18].

The proposed method will help law enforcement officials to recognize criminal persons in various places like bus and railway stations, airports, and other crowded places more efficiently. The android app developed for this research can only add images to the database and unable to delete images from the database image set of the computer. As future work, the app will be developed further to completely control the face database using an android device.

## Conflicts of Interest

The authors declare no conflicts of interest regarding the publication of this paper.